# Support Vector Machine in Prediction of Building Energy Demand Using Pseudo Dynamic Approach

*Subodh Paudel[a,b,c], Phuong H. Nguyen[b], Wil L. Kling[b], Mohamed Elmitri[c], Bruno Lacarrière[a], and Olivier Le Corre[a]\**

[a] Department of Energy Systems and Environment, Ecole des Mines de Nantes, France
[b] Department of Electrical Engineering, Eindhoven University of Technology, the Netherlands
[c] VEOLIA RECHERCHE ET INNOVATION (VERI), France
\* Corresponding Author. Tel: +33 2 51 85 82 57, Email: Olivier.Lecorre@mines-nantes.fr

**Abstract:**

Building's energy consumption prediction is a major concern in the recent years and many efforts have been achieved in order to improve the energy management of buildings. In particular, the prediction of energy consumption in building is essential for the energy operator to build an optimal operating strategy, which could be integrated to building's energy management system (BEMS).

This paper proposes a prediction model for building energy consumption using support vector machine (SVM). Data-driven model, for instance, SVM is very sensitive to the selection of training data. Thus the relevant days data selection method based on Dynamic Time Warping is used to train SVM model. In addition, to encompass thermal inertia of building, pseudo dynamic model is applied since it takes into account information of transition of energy consumption effects and occupancy profile.

Relevant days data selection and whole training data model is applied to the case studies of Ecole des Mines de Nantes, France Office building. The results showed that support vector machine based on relevant data selection method is able to predict the energy consumption of building with a high accuracy in compare to whole data training. In addition, relevant data selection method is computationally cheaper (around 8 minute training time) in contrast to whole data training (around 31 hour for weekend and 116 hour for working days) and reveals realistic control implementation for online system as well.

## 1. Introduction

The rapidly growing energy demand has drawn a significant attention in many parts of world and several countries are focusing to reduce energy consumption resulting in reduction in greenhouse gases (GHG) emission. One of the challenges foreseen nowadays is in building sector as building consumes about 40% of global energy, 25% of global water, 40% of global resources and about 1/3 of GHG emission [1]. Considering these facts, to transit into totally new sustainable building and use of efficient building material is costly and a lot of energy policies needs to be implemented to put into practice. In addition, renovation and maintenance of such building will increase overhead cost indicating adverse economies of the country. Therefore, in order to stabilize long run economy from building sector, many nations and industries, for instance, Energy Services Company (ESCOs) and Building Energy Management System (BEMS) focuses their research in energy efficiency of building. In general, energy efficiency of building depends on several phenomena such as geometrical and physical structure of building, occupant's behaviour in maintaining thermal comfort and air quality, climatic conditions and energy sources integrated to buildings.

One of the approaches to overcome barriers in energy efficiency of building is suitable demand and supply management so that by predicting energy consumption ahead, peak energy demand can be diminished and managed. Various prediction models based on physical and data-driven model (statistical, regression and artificial intelligence methods) exists today. Physical methods estimate the energy demand by using heat transfer characteristics and thermodynamics behaviour in buildings and several simulation tools TRNSYS [2], ESP-r [3], EnergyPlus [4] etc… are available. In order to reduce model equations from complex building phenomena which result in large computation time from physical methods, semi-physical methods, for example, response factor method, transfer function, frequency analysis method and lumped electrical analogy i.e. resistance and capacitance method exists [5]. However, both of these physical and semi-physical methods are highly parameterized and required several parameters of building to predict energy consumption of buildings. The computation time are also high in these methods to predict the energy demand which furthermore hurdles decision making in energy services management for ESCOs and BEMS. Other possible approaches to predict the building energy demand with limited physical parameters of the buildings are data-driven methods which rely on measurements laid down in historical databases. Statistical and regression methods seems more feasible to estimate building energy demands with limited physical information of the building, nevertheless, they are not quite accurate to include second order phenomena of building dynamics for short-term energy consumption prediction (hours to couple of days). Also, these kind of methods requires significant efforts and time to find best fitting from the actual data. In recent years, there is a growth in research work in the field of artificial intelligence like artificial neural network [6, 7, 8] and support vector machine [6, 8, 9]. These methods are accurate to explicate the complex non-linear energy consumption behaviour in buildings with limited parameters of buildings and has shown better performance than physical and statistical regression methods. In this work, support vector machine (SVM) is chosen as data driven model to predict energy consumption of building since it has higher generalization performance than neural network as it solves non-linear problems by empirical risk minimization and it always provides unique and globally optimal solution in compare to neural network which have chances of risk of local minima [9].

One of the problems of artificial intelligence methods for the prediction of building energy consumption is the necessity of proper selection of the training data since building energy consumption is governed by complex non-linear input behaviour phenomena. Thus, selection of most significant training data based on the prediction day climatic conditions and functioning profile of building is essential, which is define as relevant days data selection in this paper. With a very small data, behaviour of prediction day climatic conditions may have no similarities at all though it is computationally faster. When the training data is large, prediction day climatic conditions leads to similar and dissimilar energy consumption patterns to train the artificial intelligence model which results in learning several kind of behaviour to develop a model. This result in inflexible model for all kind of prediction as the model will be self-concentrated on the particular training sets of data to generalize and is computationally expensive as well. With the adaptability of growing the model in future, the newest environment or climatic conditions data is not considered in whole data training as its model parameter is always constant. In order to update model parameter in these whole data training to consider new datasets in future, these static learning model should be modified into online learning model. However, in relevant days data selection method, relevant training data information is changed for each day of prediction if it is relevant and thus consider newest available data in future with considering suitable training data resulting in faster computation time. Several studies have been carried to select relevant days training data based on similar trend of climatic conditions and energy load profiles for the building electricity energy consumption. For instance, various authors [10, 11] performed relevant training data selections based on climatic conditions. These methods are grounded on the Euclidean norm in the form of weighted factors to evaluate similarity between prediction day data and training data.

However, these method requires initial energy load of prediction day, which is not pragmatic for real building as this information is unavailable during prediction day. Author [12] presented Heating degree day (HDD) and Cooling degree day (CDD) to select relevant training data, however, these HDD and CDD does not precisely represent minute/hourly energy demand requirement of building as it gives average value of daily energy load. Furthermore, several authors [13, 14, 15] used clustering methods based on daily energy load to select relevant training data, nevertheless, daily average energy load during a prediction day is not practical to select because if the prediction is for couple of days, then selection of training data relies on predicted values of last couple of days and errors will be accumulated. Also, methodology applied to electricity load does not have similar behaviour to thermal i.e. heating and cooling energy consumption of building because of thermal inertia of building and thermal comfort phenomena based on internal temperature of building.

Considering all facts, selection of a relevant days training data relies mainly on single climatic conditions variable outside air temperature as a major factor in this paper since it is the strongest variable to determine heating and cooling energy consumption of buildings. Based on the fundamental understanding of building which relies on desired set-point temperature and average internal temperature of building, outside air temperature time dynamics is identified, and pattern recognition method i.e. dynamic time warping (DTW) is used to select outside air temperature similar patterns. These similar patterns are further used to identify relevant days data for particular prediction day and is used to train SVM model. Thus, model parameters of SVM is changed for each day of prediction so that it considers newest available data if it is relevant and thus adapt to different climatic conditions. The other advantages of such prediction model is that it fully relies on forecasted climatic conditions and a pseudo dynamic model which does not require previous energy load as an input to data-driven model since this model considers hidden inertial effect of building by including transition of energy consumption phenomena and occupancy profile (for details, see [16]). The paper is organized as follows: Section 2 gives an overview of the methodology with description of different steps. Section 3 highlights the case study and Section 4 discusses the results for prediction of energy consumption from DTW relevant days training data selection method and from whole data training. And, finally section 5 concludes this paper with the main highlights.

## 2. Methodology

The initial steps of methodology is the collection of building energy consumption data with several climatic variables database and approximate occupancy profile. This inputs are in the form of time series data. Block diagram of the proposed methodology for building energy consumption prediction using support vector machine is shown in Figure (1). Input to the methodology are operational energy load characteristics and dynamics of building which are further input to pseudo dynamic model. This pseudo dynamic model considers itself information about transition information of energy load characteristics, occupancy profile and thermal inertia of buildings, which is not fully dynamic but pretend to be dynamic (for details, see [16]). Temporal indicators of sample data information in a day in the form of sine and cosine form i.e. $Sin(\frac{2\pi}{L}l)$ and $Cos(\frac{2\pi}{L}l)$ are also used as an input to SVM model, where $L$ represents total number of data in one day and $l$ represents sample data of one day which varies from 1 to $L$. In addition, building database is classified based on functioning profile of occupants. The physical understanding of desired set point temperature and average internal temperature of building is used to select timing information of outside air temperature and further DTW is used to select relevant training data to train SVM model (see section 2.1). Finally, SVM model (see section 2.2) is used to predict building energy consumption based on DTW relevant days training data concept and the methodology is also compared with whole data training.

## 2.1 Relevant Days Data Selection

Relevant days data selection method means selection of similar training data for each day prediction conditions based on climatic conditions and functioning profile of building, for instance, working and weekend day in office building. Physical fundamental concept of building which relies on desired set-point temperature and average internal temperature of building is applied to select timing information of outside air temperature. Outside air temperature is considered as a major variable in this work since it has greater significance in building heating and cooling energy consumption. Energy load behaviour for Office building is shown in Figure (2) and it is clear that energy load during early morning and night is always low as there are no people in the building and energy load is higher during occupancy period. Desired set-point temperature and average internal temperature of building is also shown in Figure (2) and thus clear from Figure that the information of average internal temperature of building is unknown during prediction day. In practical, measurement of internal temperature considering multiple zone of building is rather complex and out of research, so, hypothetical physical understanding of average internal temperature of building is highlighted in Figure (2). Figure (2) illustrates the behaviour of energy load of building during prediction day and past training day (day before prediction day) and it is clear that energy load during prediction day depends on the level of average internal temperature of building. This level during prediction day, moreover, depends on the level of internal temperature of previous training days at time $t_u$. However, due to thermal dynamic behaviour of building, this average internal temperature decreases from time $t_u$ till prediction day and concludes that that energy load of prediction day depends from the steady time of average internal temperature of the day before prediction day. Consequently, this average internal temperature is highly dependent with outside air temperature and thus relevant outside air temperature is search from these time $t_u$ till prediction day from training data by using DTW.

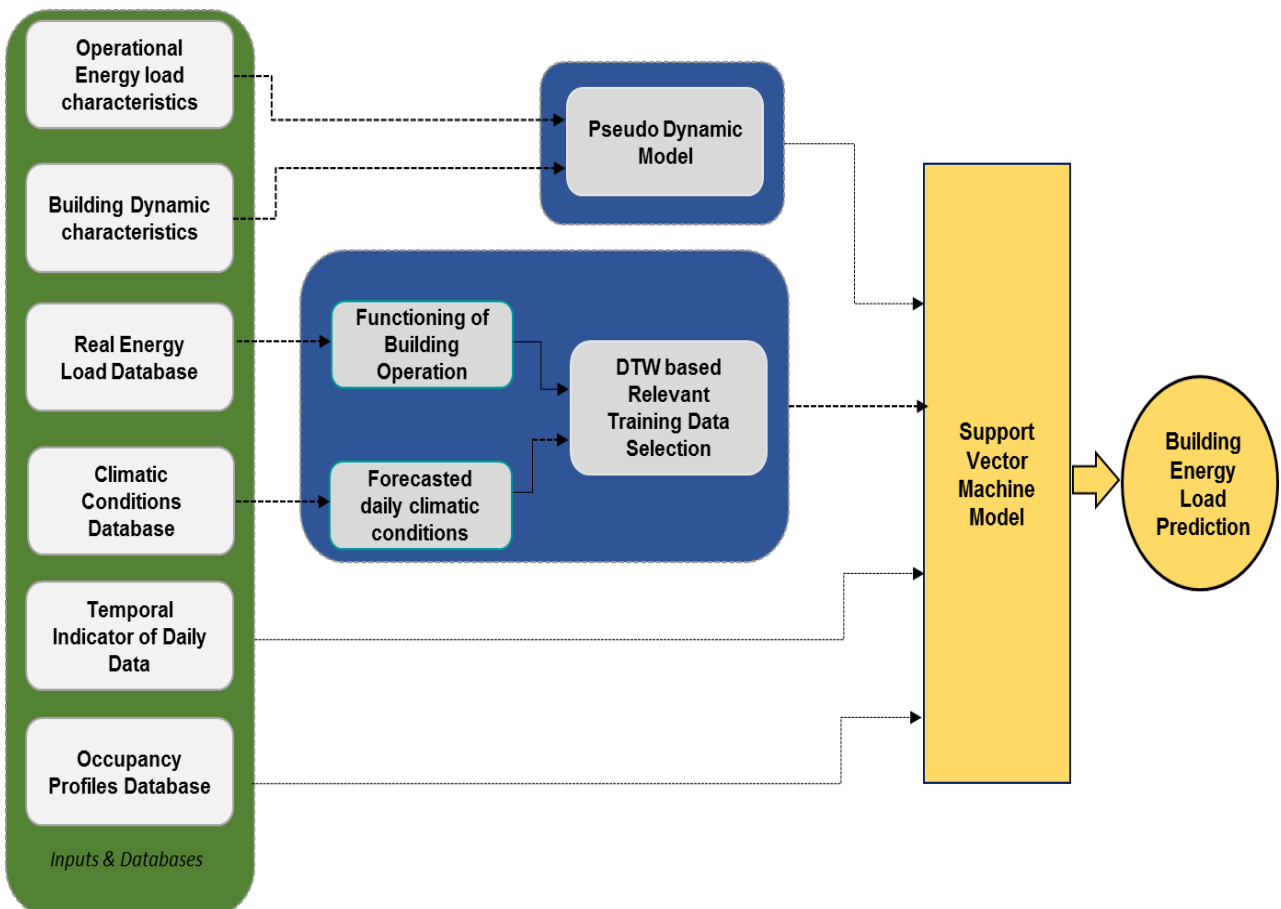

Fig. 1: Block diagram of methodology for building energy consumption prediction

DTW is a distance measure time series algorithm which identifies the similar patterns of shapes even if they are out of phase in the time axis. It relies similarity of relevant data selection based on acceleration-deceleration of signals within the time dimension. It has been widely used in science, medicine and industrial applications [17] and especially for time series prediction based on pattern recognition [18]. It has also been used to measure similarity of building energy patterns [19]. Detail about DTW presented in this work is shown in [17]. For an illustration, DTW calculates Euclidean distance measure as shown in Figure (3) between outside air temperature of training day and prediction day in two warping path. This path can be two or greater than it. In addition, path which minimizes sum of the Euclidean distance between two time series is chosen as optimal warping path. Thus, this process continues to calculate optimal Euclidean distance warping path for each day of training sample data (minute/hour) and those days which distance (weight) is low is chosen as relevant training day data selection to further train SVM model.

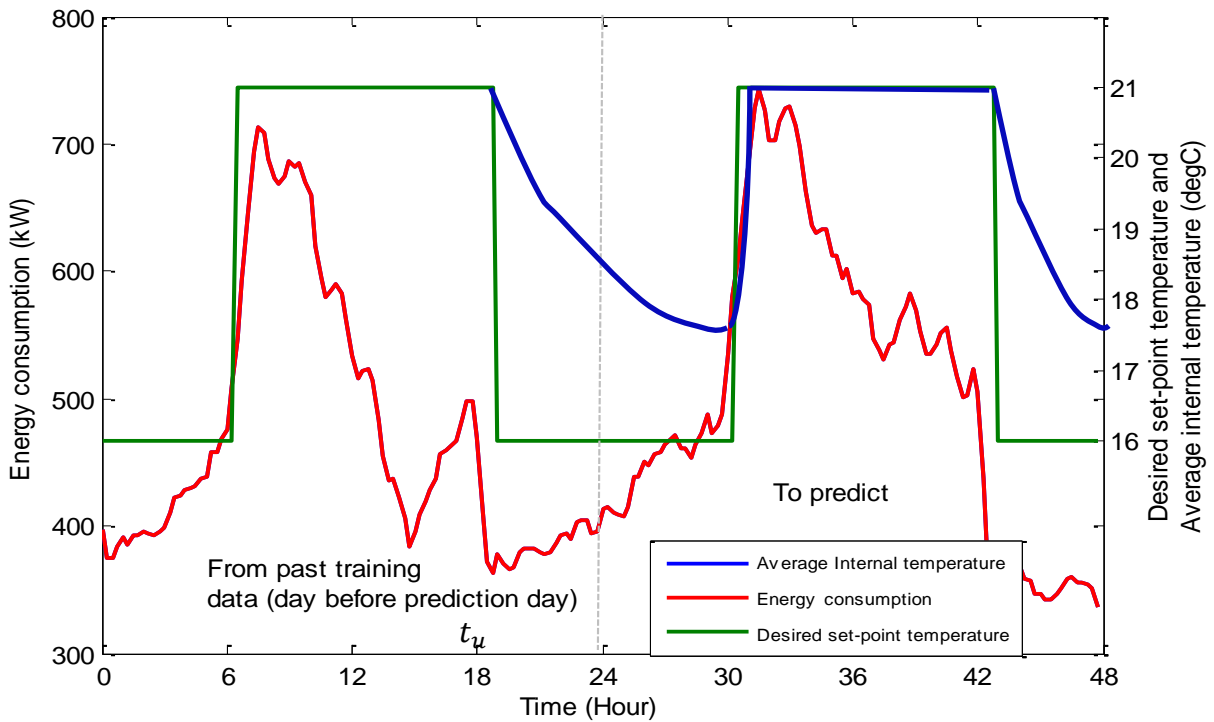

Fig. 2: Building dynamics of present and past energy load behaviour

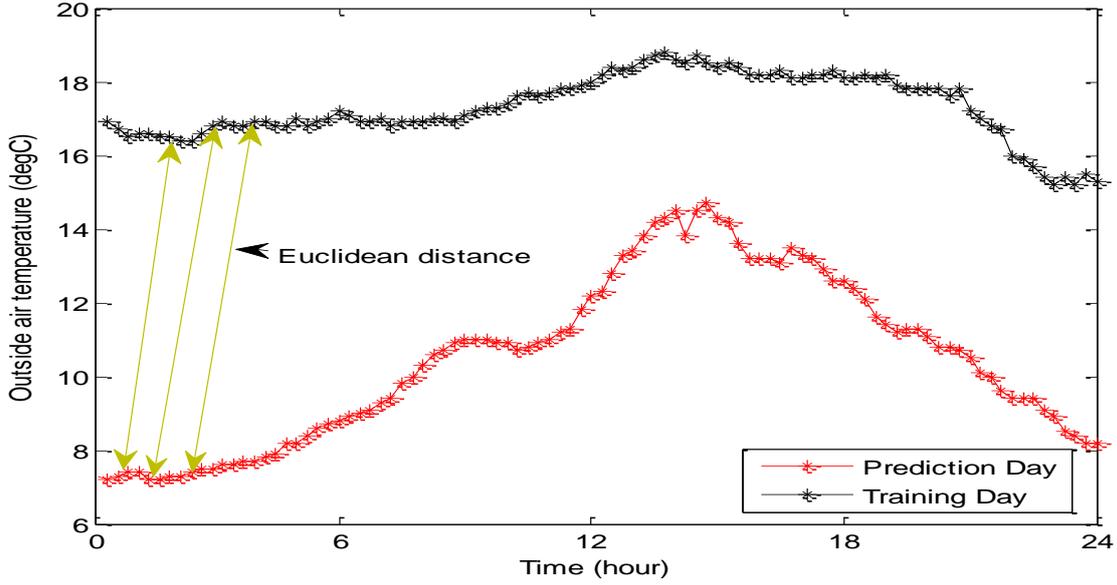

Fig. 3: DTW illustration of outside air temperature

## 2.2 Support Vector Machine

SVM is one of the most popular supervised artificial intelligence methods and widely used for classification and clustering purposes. There are several SVM available in webpage for academic and commercial purposes, such as, LibSVM [20], LS-SVMlab [21], SVMlight [22] and so forth. However, LibSVM is used in this work as it has library which could be easily integrated to Matlab, R and other programming languages interfaces with consideration for real implementation.

In general, SVM is also extended to solve regression problems, and thus support vector regression (SVR) is applied to solve non-linear regression problems by mapping non-linear regression problems to linear regression. For example, vector $x_i$ represents $i$th sample of input features (temporal indicator of sample data, climatic database, occupancy profile, operational energy load characteristics, transitional behaviour and pseudo dynamic lag) and $y_i$ represents the corresponding target value (heating energy load), then the total datasets can be represented by $\{(x_i, y_i)|_{i=1}^{n}\}$, where $x_i \in \mathbb{R}^m$ with $m$ features and $y_i \in \mathbb{R}$ and $n$ represents total number of samples in datasets. Then, SVM approximates linear relationship between input and output as shown in Equation (1).

$$f(x) = w^T x + \theta \qquad (1)$$

In Equation (1), w and θ represents weight and bias and these are estimated by minimizing regularized risk function as shown in Equation (2) [23].

$$\frac{1}{2} w^T w + C \sum_{i=1}^{n} |y_i - f(x_i)|_\varepsilon \qquad (2)$$

Where,

$$|y_i - f(x_i)|_\varepsilon = f(x) = \begin{cases} 0, & if\ |y - f(x)| < \varepsilon \\ |y_i - f(x_i)|_\varepsilon, & otherwise \end{cases}$$

In Equation (2), C is the regularization term and $|y_i - f(x_i)|_\varepsilon$ is empirical error measured by $\varepsilon$– insertion loss function. Parameter C controls trade-off between approximation error and weight vector. To estimate parameters, $\varepsilon$ insertion loss function is minimized. Equation (2), thus, illustrates that values of loss function is zero when values predicted by SVR model $f(x)$ lies within the defined tolerance level $\varepsilon$, and is magnitude of the difference between values predicted by SVR model and tolerance level $\varepsilon$ when it is outside $\varepsilon$. Furthermore, Equation (2) will again be transformed to new objective function with the introduction of slack variables, and with suitable kernel function, these objective function are optimized to estimate the parameters w and θ (for details, see [23]).

In general, there are four types of kernel function: linear, polynomial, radial basis function (RBF) and sigmoidal function. In this work, RBF kernel is used to train SVR model as it has been widely used [8,9]. RBF is a Gaussian kernel in the form $exp\left(-\gamma\|x_i - x_j\|^2\right)$ where $\gamma$ is the kernel parameter, and $x_i$ and $x_j$ are input features values for $i$th and $j$th samples of data. Thus, parameters required to estimate non-linear mapping function f($x$) are C and $\varepsilon$ for SVR, and $\gamma$ for RBF kernel.

Division of datasets is importance for generalization and in general, datasets are divided into learning/training, validation and testing in this work. In addition, normalization of input datasets are also equally importance for faster convergence. If these datasets are scaled improperly during the training/learning process, there is a risk of slower convergence in optimization problem to estimate model parameters and accuracy might decreases as well. Though there are various method of normalization, in this work, normalization with mean 0 and standard deviations 1 is performed. Finally, optimal parameters is essential for generalization in particular to smaller relevant training data since these parameters should be capable to predict with unknown data without under and over fitting problems and with high accuracy on learning/training relevant data as well. In this work, optimal SVR parameter C and $\varepsilon$, and RBF kernel $\gamma$ are estimated by using 5 fold cross validation on selected subset of relevant days training data selection.

Performance of prediction model is evaluated based on coefficient of determination ($R^2$) and root mean square (RMSE) as shown in Equation (3-4) where y is actual energy load, $\bar{y}$ is mean of actual energy load and $\hat{y}$ is predicted energy load.

$$R^2 = 1 - \frac{\sum_{i=1}^{L}(y_i - \hat{y}_i)^2}{\sum_{i=1}^{L}(y_i - \bar{y})^2} \qquad (3)$$

$$RMSE = \left(\frac{1}{L}\sum_{i=1}^{L}(y_i - \hat{y}_i)^2\right)^{\frac{1}{2}} \qquad (4)$$

## 3. Case Study

Methodology is applied to Ecole des Mines de Nantes (EMN), Office building located at Nantes, France. The building has a floor area of 25,000 m². It has 900 students and 200 employees. It consists of 120 research and administrative rooms including 30 class rooms, 3 laboratories and 8 seminar halls. Area of class room is different from each other, however, each class room can occupied 18 to 28 students. It has also 2 big and 6 seminar halls, which can accommodate up to 250 and 80 students respectively.

Building heating energy consumption data including climatic database is obtained from data acquisition system for 7 months (14/10/2012 – 28/02/2013) and (24/02/2014 – 02/05/2014) during the heating season period with 15 minutes resolution time. However, only outside air temperature of data is considered as climatic database as it has highest correlation with heating energy consumption as shown in Figure (4) (see selection of input variable and their correlations, [16]). With these database, 134 days of data from first period and one month data from second period is used for training analysis and one month remaining data from 2014 period is used for test analysis after removing outliers and missing data from measurements. First period training datasets consists high energy building data and second period training datasets consists high and low energy building data. This is because new low energy building has only operated during the second period. Outside air temperature has minimum, average and maximum temperature of -1.5$^0$C, 11.4$^0$C and 21.5$^0$C respectively. Detail about the occupancy profile, energy load characteristics during working and weekend day, transition information of energy characteristics and pseudo dynamic lag is outlined in case study of Paudel et al. [16]. From occupancy and energy load characteristics as well, building has two kinds of functioning profile: working and weekend days, thus, database is divided into two categories.

As the office building has higher energy demand during working days from Monday to Friday and less during weekend from Saturday to Sunday due to occupancy profile and considering physical fundamental understanding shown in Figure (2), DTW relevant days training data selection method relies on finding outside air temperature of last hour from 18:00 ($t_u$ in Figure 2) of previous day from prediction day till outside air temperature of prediction day during normal working days (Tuesday - Friday). For Monday, last hour of outside air temperature from 18:00 from Friday till outside air temperature of weekends (Saturday and Sunday) is searched for similar patterns. Thus, Monday depends on last two days because building will require less energy consumption to run HVAC equipment's during weekends and it needs higher energy consumption in compare to other normal working days in Monday and internal temperature of building decreases dramatically at this time because of thermal dynamic behaviour of building.

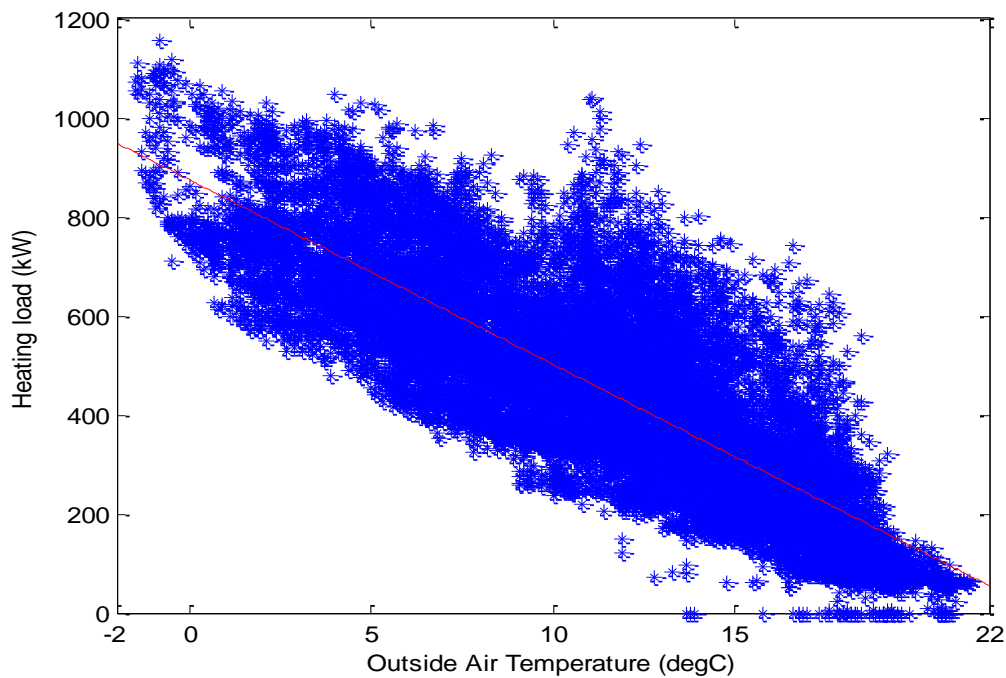

Fig.4: Correlation of Outside air temperature and heating load

Optimal number of training datasets is essential for better generalization for short-term prediction model. As there is no any robust rule for selection of number of training datasets to train any data-driven model, 12 days training data i.e. 1,152 sample datasets is selected as relevant days data since 12 days training data has higher performance in compare to other training days data (5-20) as shown in Figure (5). In the Figure (5), training datasets is increased from 5 days (480 data sample) till 20 days (1,920 data sample) and it is clear that with 12 training days datasets, the performance of prediction model is higher ($R^2$=0.96 and RMSE=21). Furthermore, this 12 days datasets are further divided into training/learning and validation into 5 fold cross validation.

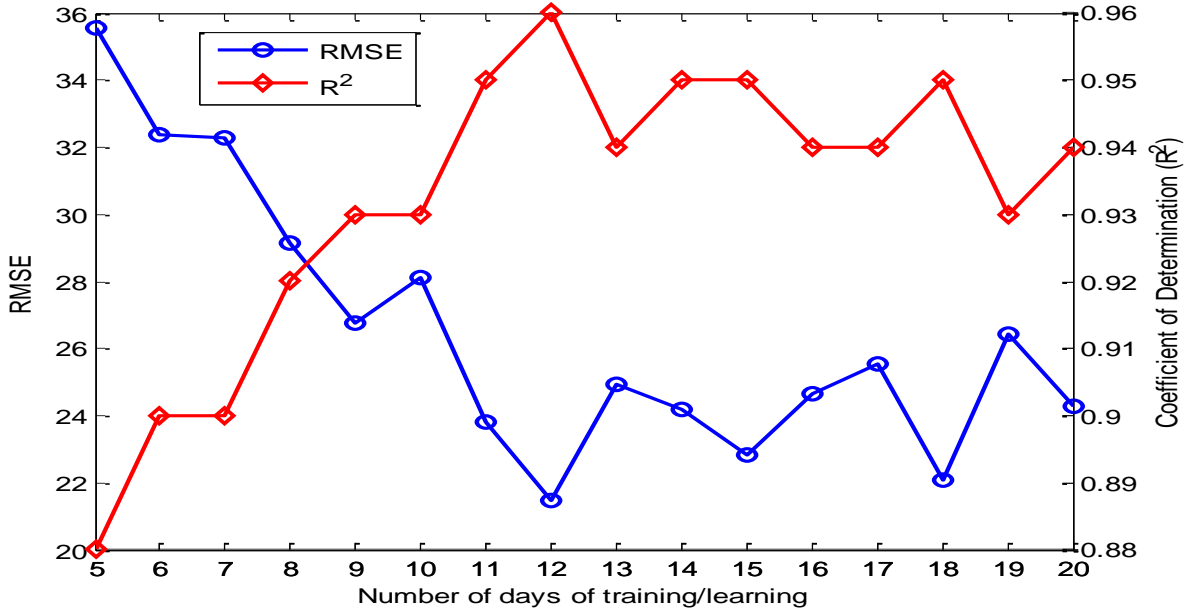

Fig. 5: Optimal number of days selection as relevant training days

In Equation (1-2), $m$ represents 9 input features: sine and cosine temporal indicator of sample data in a day, outside air temperature, energy load characteristics, transitional behaviour and four pseudo dynamic time delay, where $L$ is total number of samples data in a day i.e. 96 and $l$ varies from 1 to 96. Four pseudo dynamic time delay is used since sampling data is of 15 minute resolution and 1 hour thermal inertia is sufficient to characterize dynamics of building. Furthermore, $n$ represents 14, 016 samples of data during working day and 5,184 samples of data during weekend for whole training and 1,152 for relevant days selection method. Thus, testing data consists about 20 working days and 10 weekend. To find optimal parameters in SVR, initial searching space are $\{2^{-5}, 2^{-4}, \ldots 2^{5}\}$, $\{2^{-15}, 2^{-14}, \ldots 2^{15}\}$ and $\{0.001, 0.01, 0.1, 0.2, 0.5\}$ for C, $\gamma$ and $\varepsilon$ respectively for relevant days data selection method and $\{2^{-5}, 2^{-4}, \ldots 2^{15}\}$, $\{2^{-15}, 2^{-14}, \ldots 2^{15}\}$ and $\{0.001, 0.01, 0.1, 0.2, 0.5\}$ for C, $\gamma$ and $\varepsilon$ respectively for whole training data. The computation time is evaluated in 7*3.4Ghz CPU and 8 GB memory with windows 7 operating system.

## 4. Results and Discussion

Optimal parameters of support vector machine for each day of prediction are based on averaging validation performance ($R^2$ and RMSE) from 5 fold cross validation. Model that satisfies the minimum RMSE and maximum $R^2$ by averaging result from 5 fold cross validation will further gives optimal parameters of SVM. Thus, these optimal parameters of model will changed each day of prediction based on different training/learning datasets. The optimal parameters in DTW based relevant data training method is changed each day of prediction, however, optimal parameters C, $\gamma$ and $\varepsilon$ are 1, 8 and 0.01 for working days and 4, 8 and 0.01 for weekend in whole data training.

Figure (6) and (7) shows the prediction of energy consumption from DTW based relevant training data selection method and whole data training for working days and weekend respectively (shown only for three days). As shown in Figure (6), DTW based relevant data training has higher accuracy in prediction in compare to whole training data. It is also noticed that both model generalizes quite well approximately after 12 h till 24 h for each day prediction. However, whole training data does not generalizes quite well during the initial period (0-9 h) for each day prediction. This may be because whole data training focuses to generalize the model in terms of overall training data of outside air temperature in correlation with energy load and lacks the generality for specific hour prediction conditions, for example, during initial period (0-9 h). In contrast, DTW based relevant training has almost learn during initial period since it consider selection of training days based on these dynamic behaviour of outside air temperature and fully generalizes similar behaviour training data for particular prediction day. Accuracy and computation training time for working days and weekend in DTW based relevant training data selection and whole training for whole one month test is shown in Table (1). As shown in Table (1), $R^2$ value in relevant training was 0.10 higher than whole training data and RMSE is also lower in relevant training. In addition, computation training time for DTW based relevant data training is only 8 minute 45 sec in contrast which is 115 hour 41 min 21 sec for whole training for working days, and 7 minute 24 sec for DTW based relevant data training and 30 hour 33 minute 39 sec in whole training for weekend. Thus, with the computation time in relevant data training, it is possible to realize any practical implementation for energy demand and supply matching.

In the weekend as well, DTW based relevant training method is superior to whole training for prediction. It is clear that both method have higher RMSE in compare to working day. The reason behind this might be testing data consists new additional low energy building data which has only few training datasets fall in this period which result in poor generalization in whole training. However, computation time in relevant data training during weekend is 7 min 24 sec which is lower than relevant data training for working days. This is because of time DTW searches for best training data from 5,184 training samples in weekend in compare to 14, 016 training samples in working day.

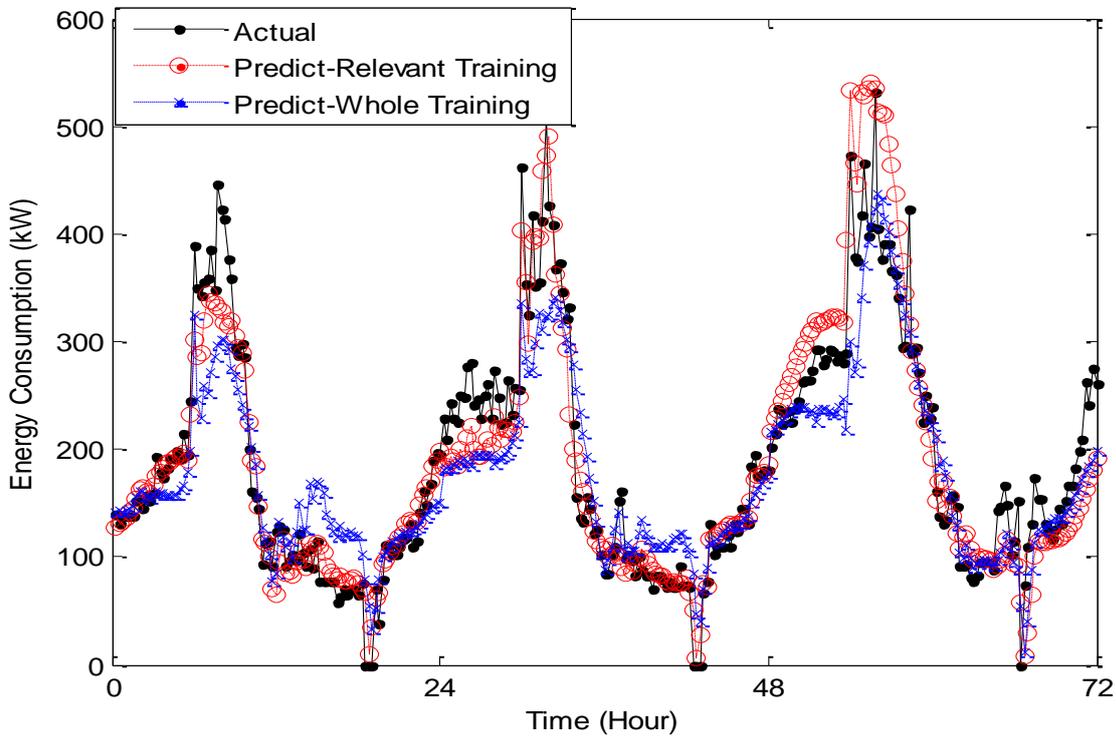

Fig. 6: Prediction of heating energy consumption for working days

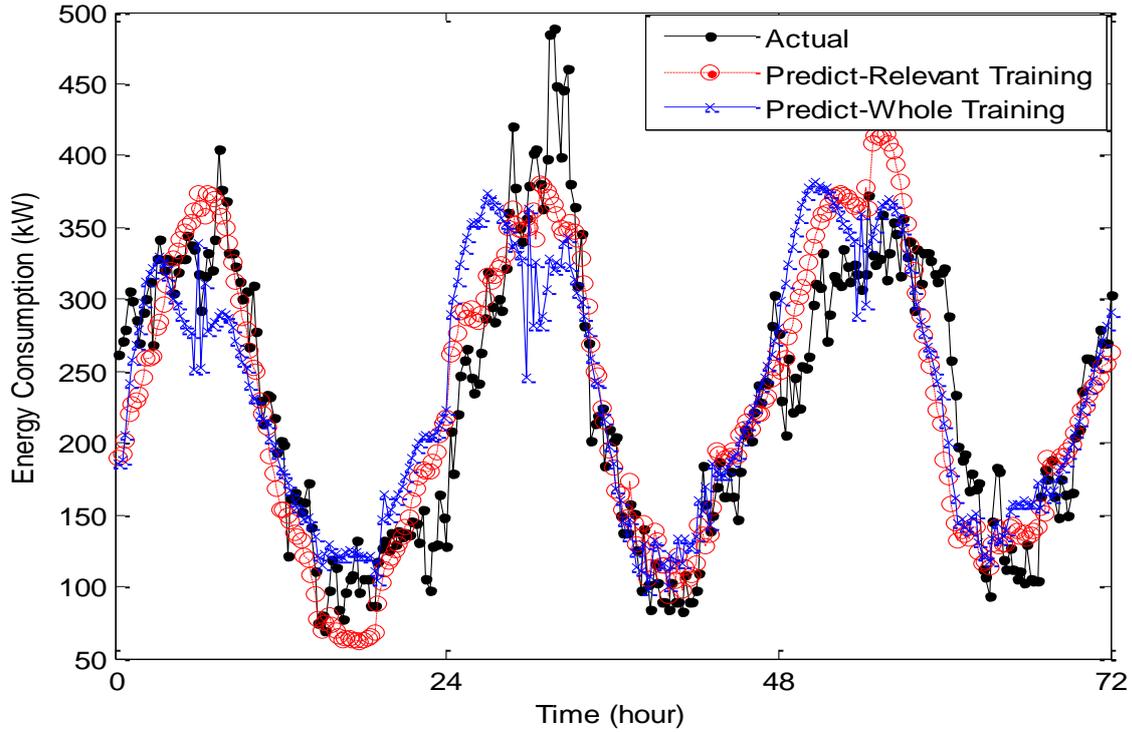

Fig. 7: Prediction of heating energy consumption during weekend

Overall, it is clear that training/learning data has significant role in accuracy of prediction of building energy consumption and due to similarities and dissimilarities data present in whole data training, performance of whole data training is lower than DTW based relevant training method.

Table 1: Performance and computation time of DTW based relevant days data selection method and whole data training in prediction of energy consumption during working days/weekend

| Performance Measure | Working Day | | Weekend | |
|---|---|---|---|---|
| | Relevant training | Whole training | Relevant training | Whole training |
| $R^2$ | 0.88 | 0.76 | 0.82 | 0.69 |
| RMSE | 51 | 73 | 50 | 140 |
| Computation training time | 8 min 45 sec | 115h 41 min 21sec | 7 min 24 sec | 30h 33 min 39sec |

## 5. Conclusion

This paper thus predict building energy consumption using DTW based relevant days training data selection method and compare with whole data training. Outside air temperature is taken as a major variable to select relevant days training data selection as this variable is strongest in determining heating and cooling energy consumption of buildings. Physical understanding based on average internal temperature and desired set-point temperature of building is used to select outside air temperature time dynamics. DTW is further used to select similar patterns of outside air temperature. Result showed that DTW based on relevant data selection has higher accuracy ($R^2$=0.88, RMSE=51) in compare to ($R^2$=0.76, RMSE=73) for whole training during working days and ($R^2$=0.82, RMSE=50) for relevant training and ($R^2$=0.69, RMSE=140) for during weekend. In addition, computation time is too short in relevant days data selection method in compare to whole data training.

## Acknowledgement

This research has been done in collaboration with Ecole des Mines de Nantes, Eindhoven University of Technology and VEOLIA RECHERCHE ET INNOVATION (VERI), funded through Erasmus Mundus Joint doctoral programme SELECT+, the support of which is gratefully acknowledged.

## References

1. UNEP. 2013. Energy Efficiency for Buildings. http://www.studiocollantin.eu/pdf/UNEP%20Info%20sheet%20-%20EE%20Buildings.pdf
2. TRNSYS 17, a TRaNsient System Simulation program. http://sel.me.wisc.edu/trnsys/features
3. Citherlet, S., 2001. Towards the holistic assessment of building performance based on integrated simulation approach. PhD Thesis, Swiss Federal Institute of Technology. (http://www.esru.strath.ac.uk/Documents/PhD/citherlet_thesis.pdf)
4. Crawley, D.B., Lawrie, L.K., Winkelmann, F.C., Buhl, W.F., Huang, Y.J., Pedersen, C.O., Strand, R.K., Liesen, R.J., Fisher, D.E., Witte, M.J., Glazer, J., 2001. EnergyPlus: creating a new-generation building energy simulation program, Energy and Buildings 33, 319-331.
5. Underwood, C.P., Yik, F.W.H., 2004. Modeling methods for energy in buildings, Blackwell Science.
6. Zhao, H., Magoules, F., 2012. A review on the prediction of building energy consumption. Renewable and Sustainable Energy Reviews 16, 3586-3592.
7. Yokoyama, R., Wakui, T., Satake, R., 2009. Prediction of energy demands using neural network with model identification by global optimization. Energy Conversion and Management 50, 319-327.
8. Li, Q., Meng, Q., Cai, J., Yoshino, H., Mochida, A., 2009. Predicting hourly cooling load in the building: A comparision of support vector machine and different artificial neural networks. Energy Conversion and Management 50, 90-96.
9. Dong, B., Cao, C., Lee, S.E., 2005. Applying support vector machines to predict building energy consumption in tropical region. Energy and Buildings 37, 545-553.
10. Mandal, P., Senjyu, T., Urasaki, N., Funabashi, T., 2006. A neural network based several-hour ahead electrical load forecasting using similar days approach. Electrical Power and Energy Systems 28, 367-373.
11. He, Y-J., Zhu, Y-C., Gu, J-C., Yin, C-Q., 2005. Similar day selecting based neural network model and its application in short term load forecasting. Proceedings of the Fourth International Conference on Machine Learning and Cybernetics, IEEE, 18-21 August, Guangzhou, China.
12. Roldan-Blay, C., Escriva-Escriva, G., Alvarez-Bel, C., Roldan-Porta, C., 2013. Upgrade of an artificial neural network prediction method for electrical consumption of forecasting using an hourly temperature curve model. Energy and Buildings 60, 38-46.
13. Ghanbari, A., Ghaderi, S.F., Azadeh, M.A., 2010. A clustering based genetic fuzzy expert system for electrical energy demand prediction. Second International Conference on Computer and Automation Engineering, 26-28 February, Singapore, pp. 470-411.

14. Duan, D-X., 2009. Short-term load prediction based on ant colony clustering elman neural network model. Second International Workshop on Computer Science and Engineering, 28-30 October, Qingdao, China, pp. 394-397.
15. Grzenda, M., Macukow, B., 2006. Demand prediction with multi-stage neural processing. Advances in Natural Computation and Data Mining, pp. 131 – 141.
16. Paudel, S., Elmtiri, M., Kling, W.L., Lacarrière, B., Le Corre, O., 2014. Pseudo dynamic transitional modeling of building heating energy demand using artificial neural network. Energy and Buildings 70, 81-93.
17. Keogh, E., Ratanamahatana, C.A., 2005. Exact indexing of dynamic time warping. Knowledge and Information Systems 7, 358-386.
18. Mager, J., Paasche, U., Sick, B., 2008. Forecasting financial time series with support vector machines based on dynamic kernels. IEEE Conference on Soft Computing in Industrial Applications, 25-27 June, Muroran, Japan, pp. 252-257.
19. Iglesias, F., Kastner, W., 2013. Analysis of similarity measures in time series clustering for the discovery of building energy patterns. Energies 6, 579-597. doi:10.3390/en6020579
20. Chang, C.-C., Lin, C.-J., 2001. LIBSVM. A library for support vector machines. Technical Report. Department of Computer Science and Information Engineering, National Taiwan University, Taiwan. http://www.csie.ntu.edu.tw/~cjlin/libsvm/
21. Pelckmans, K., Suykens, J.A.K., Vangestel, T., De Brabanter, J., Lukas, L., Hamers, B., De Moor, B., Vandewalle, J., 2002. LS SVMlab: a MATLAB/c toolbox for least square support vector machines. Tutorial. KU Leuven-ESAT, Leuven.
22. Joachims, T., 1999. Making large-scale SVM learning practical. Advances in kernel methods: support vector learning, pp. 169-184.
23. Vapnik, V.N., 1995. The nature of statistical learning theory. Springer-Verlag New York, Inc., New York, NY, USA.